\documentclass{article}
\usepackage[dblblindworkshop,final]{neurips_2026}
\usepackage[utf8]{inputenc}
\usepackage[T1]{fontenc}
\usepackage{graphicx}
\usepackage{amsmath}
\usepackage{amssymb}
\usepackage{booktabs}
\usepackage{hyperref}
\usepackage{url}
\usepackage{nicefrac}
\usepackage{microtype}
\usepackage{xcolor}
\DeclareMathOperator*{\argmax}{arg\,max}

\title{How Perturbations Propagate: A Multi-Level Analysis of Robustness in Large Language Models}
\workshoptitle{Interpretability as a Science}

\author{
Dun Li Chan$^{1}$ \quad
Emily Liu$^{2}$\thanks{Corresponding author: \texttt{emily.zf.liu@gmail.com}} \quad
Niyathi Allu$^{2}$ \quad
Christian Hoang$^{3}$ \\
$^{1}$INTI International College Penang \\
$^{2}$Independent Researcher \\
$^{3}$FPT University \\
}

\begin{document}

\maketitle

\begin{abstract}
{Language models encounter typos, corrupted text, altered words, and disrupted token order, yet robustness is usually evaluated only through output behavior. We study how six naturalistic and synthetic input perturbations propagate through decoder-only language models at three levels: output behavior, hidden-state geometry, and attention-head function. We evaluate behavioral effects across four GPT-2 and two Qwen2.5 checkpoints, analyze layerwise geometry using centered kernel alignment and intrinsic dimension, and examine attention-head responses in GPT-2. Perturbation types produce distinguishable metric profiles that are not fully captured by output measures and are only partly consistent across the tested checkpoints. Copying scores show the strongest pooled associations with activation-patching recovery under token substitution and shuffling, although these associations do not isolate copying-specific effects. Gradient-guided HotFlip perturbations also cause stronger behavioral and representational disruption than rate-matched random token substitutions in GPT-2; their behavioral effects are consistent across all six tested checkpoints. Our results show that robustness claims based on a single behavioral or representational metric can be misleading, and motivate multi-level evaluation of how perturbations alter language-model computation.}
\end{abstract}

\section{Introduction}

Language models are often evaluated on clean and well-formed text, but deployed inputs usually contain typographical errors or character corruption from optical character recognition, which alter wording and disrupt token order. Prior work has shown that such changes can substantially degrade language-model behaviour even when they preserve much of the text's meaning for a human reader \citep{belinkov2018synthetic,pruthi2019combating}. Adversarial attacks make this vulnerability more explicit by selecting small input edits that optimise a model-level objective \citep{ebrahimi2018hotflip,wallace2019universal,jin2020textfooler}. However, output degradation alone does not explain how a perturbation changes the computation performed inside the model.

Two perturbations can yield similar changes in generated text while disrupting different representations or computational components. Conversely, a small output change may conceal a substantial shift in hidden-state geometry. Existing robustness evaluations therefore provide limited evidence about whether perturbation effects are shared across corruption types, localize to recognizable mechanisms, or generalize across model scales and families.

Thus, we address these questions through a multi-level analysis of perturbation propagation in decoder-only language models. We compare six naturalistic and synthetic perturbation types (character substitution, keyboard typos, random token substitution, token shuffling, word substitution, and synonym substitution) using WikiText-2 inputs. We measure output behavior using negative log-likelihood and generated-output divergence across GPT-2 models of increasing scale and Qwen2.5 models; representation change using CKA \citep{kornblith2019similarity} and intrinsic-dimension shifts \citep{facco2017estimating,aghajanyan2021intrinsic}; and component-level responses using attention-head function scores and activation patching \citep{olsson2022context,wang2022interpretability,heimersheim2024patching}. We also compare gradient-guided HotFlip perturbations against random token substitutions matched at the accepted edited positions.

Our contributions are:

\begin{itemize}
    \item We provide a multi-level empirical analysis of six text perturbation types, jointly measuring output behavior, representation geometry, and attention-head responses.
    \item We test whether perturbation effects are associated with functionally characterized attention heads using attention-based measurements and activation patching.
    \item We evaluate the stability of perturbation signatures across GPT-2 scale and the Qwen2.5 model family.
    \item We compare adversarial and rate-matched random token substitutions, separating adversarial optimization effects from edit rate alone.
\end{itemize}

\section{Background and Related Work}

\paragraph{Input perturbation and adversarial robustness in NLP.}
Character- and word-level perturbations, including typographical errors and noise introduced by imperfect text extraction, can substantially degrade model performance despite remaining easy for humans to interpret \citep{belinkov2018synthetic,pruthi2019combating}. A related literature constructs adversarial perturbations optimized against model behavior, including gradient-guided token substitutions and universal triggers \citep{ebrahimi2018hotflip,wallace2019universal,jin2020textfooler}. These studies primarily evaluate output behavior or task performance. We instead ask whether perturbation types leave distinguishable \emph{internal} signatures, and whether an adversarially optimized perturbation is internally distinguishable from a naturalistic perturbation with matched position and rate.

\paragraph{Representational similarity metrics.}
Centered Kernel Alignment (CKA) is widely used to compare neural-network representations \citep{kornblith2019similarity}. However, CKA can be sensitive to outlier dimensions and simple transformations that do not necessarily correspond to meaningful functional differences \citep{davari2023reliability}. We therefore use CKA as one diagnostic among several, rather than interpreting it as a complete measure of internal disruption.

\paragraph{Intrinsic dimension of learned representations.}
High-dimensional representations may lie near lower-dimensional manifolds. Nearest-neighbor estimators such as TwoNN use local distance scaling to estimate this intrinsic dimension \citep{facco2017estimating}. Recent work has applied intrinsic-dimension analyses to language models to study fine-tuning, representation geometry, truthfulness, and training dynamics \citep{aghajanyan2021intrinsic,yin2024truthfulness,razzhigaev2024shape,ruppik2025less}. We extend this line of work by treating the \emph{change} in intrinsic dimension between clean and perturbed inputs as a perturbation-response metric.

\paragraph{Attention-head function and mechanistic interpretability.}
Mechanistic interpretability research has identified recurring attention-head functions, including induction heads that support sequence continuation \citep{elhage2021mathematical,olsson2022context} and copying or name-mover heads whose output-value circuits promote attended tokens in the output distribution \citep{wang2022interpretability}. These functions are typically studied on clean, hand-constructed tasks. We test how head-function measures and perturbation responses relate across naturally occurring and synthetic input corruptions.

\paragraph{Activation patching methodology.}
Activation patching is a widely used intervention for testing whether internal activations contribute to a model behavior \citep{meng2022locating,heimersheim2024patching}. Its conclusions depend on the corruption procedure, metric, and alignment between clean and corrupted runs \citep{heimersheim2024patching}. We therefore restrict patching analyses to perturbations that preserve token count after retokenization and interpret recovery as evidence of functional association within this experimental setting.

\section{Methods}

\subsection{Inputs and perturbations}

We use the WikiText-2 Raw dataset \citep{merity2016pointer} as the source of clean input sequences. We retain sequences of length at least 128 and randomly sample 300 sequences using a fixed seed for each experiment. This fixed evaluation set is used across perturbation types and models.

We evaluate the Hugging Face checkpoints \texttt{openai-community/gpt2}, \texttt{gpt2-medium}, \texttt{gpt2-large}, and \texttt{gpt2-xl} \citep{radford2019language}, together with \texttt{Qwen/Qwen2.5-0.5B} and \texttt{Qwen/Qwen2.5-1.5B} \citep{qwen2025qwen25}. Behavioral comparisons use all six checkpoints; the attention-head analysis and full adversarial representation comparison use GPT-2.

For a clean text sequence \(x\), we construct a perturbed sequence \(\tilde{x}\) at perturbation strength \(p \in [0,1)\). We consider six perturbation types spanning character-, word-, and token-level changes.

\paragraph{Character perturbations.}
Under uniform character substitution (\texttt{char}), each character is independently replaced with probability \(p\) by a character drawn uniformly from digits, upper- and lower-case letters, and common punctuation. Under typographical noise (\texttt{typo}), selected characters are replaced by adjacent keys on a QWERTY keyboard, approximating common typing errors.

\paragraph{Word and token substitutions.}
Under uniform token substitution (\texttt{token}), each token is independently replaced with probability \(p\) by a token sampled uniformly from the model vocabulary. Under uniform word substitution (\texttt{word}), the replaced unit is a whitespace-delimited word and the replacement is sampled from alphabetic vocabulary tokens. Under synonym substitution (\texttt{synonym}), selected words are replaced using synonyms from WordNet \citep{miller1995wordnet}.

\paragraph{Token shuffling.}
Under token shuffling (\texttt{shuffle}), we select a contiguous window containing approximately \(pN\) tokens, where \(N\) is the sequence length, and randomly permute the tokens within that window. Tokens outside the window are unchanged. This intervention preserves token identity while disrupting local order and syntax.

\subsection{Behavioral, representational, and head-level measurements}

\paragraph{Output metrics.}
We evaluate output-level effects using negative log-likelihood (NLL) and normalized Levenshtein distance between generated clean and perturbed outputs, which we report as output divergence. Definitions are given in Appendix~\ref{output_metrics}.

\paragraph{Representation metrics.}\label{cka}
We compare clean and perturbed hidden states with centered kernel alignment (CKA) \citep{kornblith2019similarity}. For layer \(l\), let \(a\) and \(\tilde{a}\) denote the clean and perturbed activation matrices, and let \(H=I_n-\frac{1}{n}\mathbf{1}_n\mathbf{1}_n^\intercal\) be the centering matrix. With \(a_c=Ha\) and \(\tilde{a}_c=H\tilde{a}\), we compute
\begin{align*}
\mathrm{CKA}_l(a,\tilde{a}) =
\frac{\left\|a_c^\intercal\tilde{a}_c\right\|_F^2}
{\left\|a_c^\intercal a_c\right\|_F
\left\|\tilde{a}_c^\intercal\tilde{a}_c\right\|_F}.
\end{align*}
We evaluate layers \(l\in\{1,\ldots,n_{\mathrm{layer}}\}\), excluding layer 0 because it directly reflects the changed input embedding. CKA requires activation matrices of equal shape; we therefore filter sequences shorter than 128 and truncate inputs to length 128. Token substitution and shuffling preserve token count exactly, whereas character, typo, word, and synonym perturbations can alter tokenization. For the latter perturbations, truncation restores equal matrix size but not semantic token correspondence. Their CKA values therefore reflect the combined effects of retokenization and representation change rather than a strictly position-aligned comparison. To reduce sensitivity to high-variance outlier dimensions \citep{davari2023reliability}, we rank dimensions by their variance in the clean activation matrix and remove the five highest-variance dimensions from both the clean and perturbed matrices before computing CKA.

We additionally measure the intrinsic dimension of layerwise activations using the two-nearest-neighbors (TwoNN) estimator \citep{facco2017estimating}. Intrinsic dimension provides a local measure of representational complexity: perturbations that disrupt regular structure may change the effective dimension of the activation manifold. We report the change in intrinsic dimension between clean and perturbed activations; the estimator is defined in Appendix~\ref{intrinsic_dim_metrics}.

\paragraph{Head-function scores.}
To characterize the functional roles of attention heads, we compute four head-level scores. The previous-token score measures attention to the immediately preceding token; the duplicate score measures attention to earlier occurrences of the current local subsequence; and the induction score measures attention from a repeated prefix to the token that followed its earlier occurrence. We also compute a copying score based on the head's OV circuit \citep{wang2022interpretability}. For the copying probe, with head-specific value matrix \(W_v^{(h)}\), output projection \(W_o^{(h)}\), token embedding matrix \(E\), and language-model head \(U\), we define
\begin{align*}
C = E W_v^{(h)}W_o^{(h)}U,
\qquad C\in\mathbb{R}^{|V|\times|V|}.
\end{align*}
Letting \(C_i\) denote row \(i\) of \(C\), the copying score is
\begin{align*}
\mathrm{CopyingScore}(h)
=
\mathbb{E}_{i\in[V]}
\left[
\mathbf{1}
\left[
i\in\mathrm{top}_k\text{-}\mathrm{indices}(C_i)
\right]
\right].
\end{align*}
The copying score is estimated from 5,000 randomly sampled vocabulary token embeddings passed through the head's OV circuit and unembedding, with $k=5$. This probe does not use the contextual next-token targets of the evaluation sequences. Its top-$k$ criterion is invariant to uniform positive scaling of the OV circuit. For token substitution and shuffling, we use corrected head-level scores from the same probe implementation; the other four entropy conditions retain their original scores. Definitions of the previous-token, duplicate, and induction scores are given in Appendix~\ref{attn_func_scores}.

\paragraph{Perturbation responses of attention heads.}\label{patchingdesc}
We examine how functional head types respond to perturbation in two ways. First, we compute the change in normalized attention entropy at 30\% perturbation, which measures whether a head's attention distribution becomes more diffuse or more concentrated. The entropy definition is provided in Appendix~\ref{attnentnormdelta}.

Second, we use clean activation patching: for clean input \(x\) and perturbed input \(\tilde{x}\), we replace the perturbed output activation of head \(h\) with its corresponding clean activation and obtain patched output \(y_p\). For an output metric \(O\in\{\mathrm{NLL},\mathrm{OutputDivergence}\}\), we report recovery as
\begin{align*}
\Delta\%O =
\frac{O(y_p)-O(\tilde{y})}
{O(y)-O(\tilde{y})}
\times100,
\end{align*}
where \(y\) and \(\tilde{y}\) are the outputs on clean and perturbed inputs. Because activation patching requires positional correspondence, we restrict this analysis to token substitution and shuffling perturbations.

\paragraph{Head-level association controls.}
Tables~\ref{tab:attention_entropy}--\ref{tab:patching} retain descriptive pooled correlations. For a separate analysis, we average observations within each layer--head pair and compute correlations across distinct heads. Layerwise plots correlate the 12 heads within each GPT-2 attention layer. We also evaluate the supplied corrected copying score's recovery association across 144 head means after accounting for layer indicators and the supplied head-level OV-circuit norm. We rank copying score, recovery, and OV norm, regress the first two ranked variables separately on an intercept, layer indicators, and ranked OV norm, and correlate the residuals. Appendix~\ref{headcontrols} reports these partial rank correlations and exploratory head-bootstrap intervals.

\subsection{Adversarial comparison}

The non-adversarial perturbations above need not approximate a worst-case input change. We therefore compare random token substitution with a HotFlip-style gradient-guided token attack \citep{ebrahimi2018hotflip} that maximizes sequence-level NLL.

For perturbation strength \(p\), we select \(\lfloor pN\rfloor\) token positions uniformly at random, where \(N\) is the sequence length. At each selected position \(i\), we compute the gradient of the loss with respect to that token's embedding and construct a shortlist \(S\) of the 50 vocabulary tokens with the largest first-order estimated loss increase. We then evaluate shortlisted replacements directly and set the token to
\begin{align*}
\argmax_{t\in S\cup\{x_i^{\prime 0}\}}
\mathrm{NLL}\!\left(f\!\left(x'\mid_{x_i'=t}\right)\right),
\end{align*}
where \(f\) is the model, \(x'\mid_{x_i'=t}\) denotes the current adversarial sequence with token \(i\) set to \(t\), and \(x_i^{\prime0}\) is the original token at that position. A replacement is committed only when it increases NLL; hence, \(\lfloor pN\rfloor\) is an upper bound on the number of accepted edits. Shortlist hyperparameters are reported in Appendix~\ref{advshortlist}.

For every accepted adversarial substitution, we construct a paired random-token control that changes the same position using a uniformly sampled vocabulary token. This control matches the adversarial attack in both perturbation type and realized edited positions. We compare adversarial and rate-matched random substitutions using the same output and representation metrics used for the naturalistic perturbations: NLL, output divergence, CKA, and intrinsic-dimension change.

\section{Experiments}

\subsection{RQ1: Perturbation propagation across levels of representation}\label{rq1}

\begin{table}[t]
\centering
\small
\renewcommand{\arraystretch}{1.12}
\setlength{\tabcolsep}{3.2pt}

\caption{GPT-2 output divergence and negative log-likelihood (NLL) under six perturbation types. Entries are mean$_{\mathrm{SD}}$ across evaluation sequences. Perturbation rate is reported as a percentage.}
\label{tab:ptb_compare_behavioral}

\vspace{2pt}
\begin{tabular}{lrrrrrr}
\toprule
Perturbation rate & Char. & Typo & Token & Word & Synonym & Shuffle \\
\midrule
\multicolumn{7}{l}{\textit{Output divergence}} \\
0\%  & $0.000_{0.000}$ & $0.000_{0.000}$ & $0.000_{0.000}$ & $0.000_{0.000}$ & $0.000_{0.000}$ & $0.000_{0.000}$ \\
5\%  & $0.544_{0.075}$ & $0.514_{0.068}$ & $0.179_{0.088}$ & $0.205_{0.064}$ & $0.130_{0.057}$ & $0.111_{0.145}$ \\
30\% & $0.883_{0.025}$ & $0.796_{0.032}$ & $0.577_{0.060}$ & $0.569_{0.043}$ & $0.421_{0.076}$ & $0.387_{0.164}$ \\
50\% & $0.932_{0.015}$ & $0.852_{0.029}$ & $0.669_{0.037}$ & $0.661_{0.031}$ & $0.543_{0.067}$ & $0.507_{0.128}$ \\
\addlinespace[3pt]
\midrule
\multicolumn{7}{l}{\textit{Negative log-likelihood}} \\
0\%  & $3.917_{0.383}$ & $3.917_{0.383}$ & $3.917_{0.383}$ & $3.917_{0.383}$ & $3.917_{0.383}$ & $3.917_{0.383}$ \\
5\%  & $6.096_{0.437}$ & $5.908_{0.427}$ & $4.806_{0.577}$ & $4.674_{0.410}$ & $4.212_{0.411}$ & $4.372_{0.774}$ \\
30\% & $7.213_{0.208}$ & $7.194_{0.264}$ & $8.569_{0.688}$ & $7.376_{0.388}$ & $5.427_{0.434}$ & $6.016_{1.082}$ \\
50\% & $6.732_{0.193}$ & $6.963_{0.264}$ & $10.238_{0.520}$ & $8.561_{0.315}$ & $6.069_{0.463}$ & $6.929_{1.008}$ \\
\bottomrule
\end{tabular}
\end{table}

\begin{figure}[t]
    \centering
    \includegraphics[width=\linewidth]{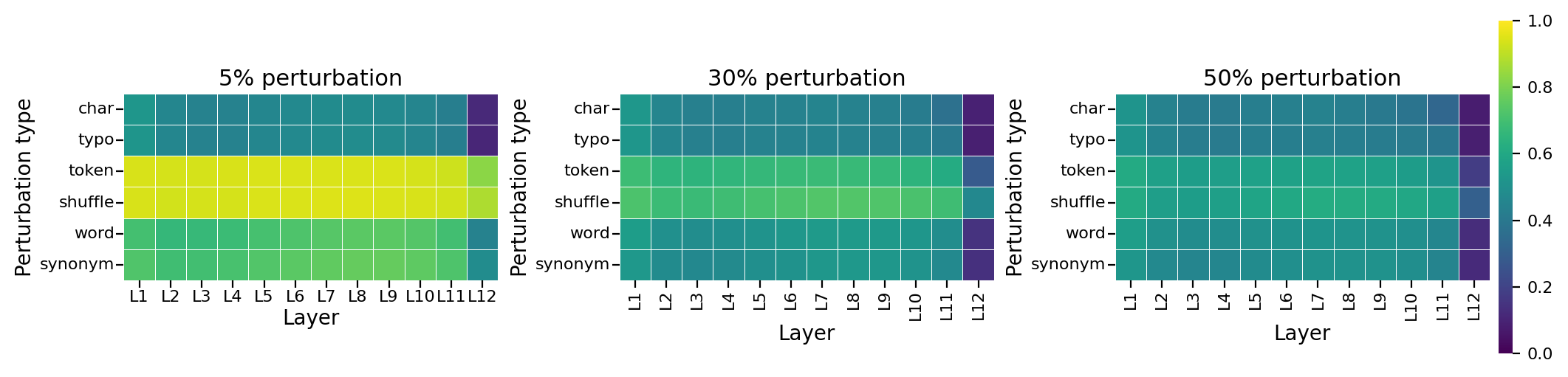}
    \caption{Layerwise representational similarity between clean and perturbed GPT-2 activations, measured by centered kernel alignment (CKA). Rows show the six perturbation types and columns show transformer layers. The three panels correspond to perturbation rates of 5\%, 30\%, and 50\%. Higher values indicate greater similarity to the clean representation. For character, typo, word, and synonym perturbations, values combine representation change with retokenization and positional mismatch.}
    \label{fig:cka-heatmaps}
\end{figure}

\begin{figure}[t]
    \centering
    \includegraphics[width=\linewidth]{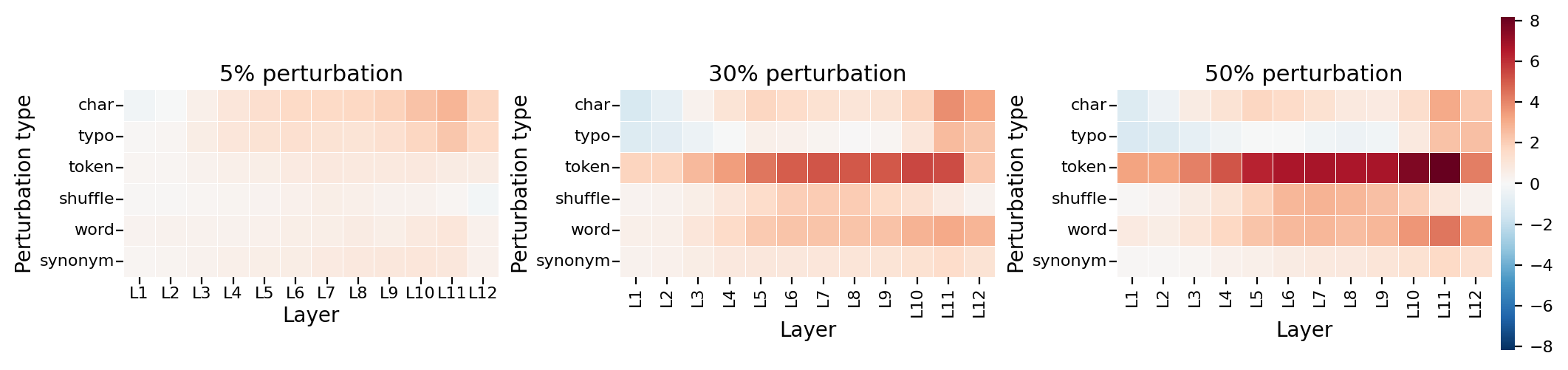}
    \caption{Layerwise change in the local intrinsic dimension of GPT-2 representations under perturbation, estimated using TwoNN. Positive values indicate an increase relative to the corresponding clean representation. Rows show perturbation types, columns show transformer layers, and the three panels correspond to perturbation rates of 5\%, 30\%, and 50\%.}
    \label{fig:intrinsic-dimension-heatmaps}
\end{figure}

Perturbation types have distinct behavioural and internal profiles. In Table~\ref{tab:ptb_compare_behavioral}, character substitution produces the largest output divergence, while keyboard typos are less disruptive. Token and word substitutions have similar output divergence but substantially different NLL, showing that generated-text change and predictive confidence need not agree. Shuffling changes both metrics more gradually than substitution-based corruptions.

The representation metrics provide a related but non-identical ordering. CKA generally decreases with perturbation strength (Figure~\ref{fig:cka-heatmaps}); character and typo perturbations yield the lowest similarity, while shuffling retains comparatively high similarity. For character, typo, word, and synonym perturbations, however, retokenization can change token correspondence: these CKA differences cannot be attributed solely to changes at matched semantic positions. TwoNN estimates also distinguish the perturbations (Figure~\ref{fig:intrinsic-dimension-heatmaps}): at higher rates, character and typo corruption lower estimated intrinsic dimension in early layers, whereas token substitution yields the largest positive change. These are descriptive geometric profiles, with alignment-dependent comparisons requiring additional controls.

\subsection{RQ2: Attention head function under perturbation}
\begin{table}[t]
\centering
\small
\renewcommand{\arraystretch}{1.15}
\setlength{\tabcolsep}{4.5pt}

\caption{Spearman correlations between head-function scores and changes in normalized attention entropy under 30\% perturbation on GPT-2. Entropy change is perturbed minus clean. These descriptive correlations pool observations from all 144 GPT-2 heads; pooled uncertainty intervals are not reported. Copying entries use supplied corrected scores for token and shuffle, and original scores for the other four conditions. Layerwise head-mean analyses are given in Appendix~\ref{headcontrols}.}
\label{tab:attention_entropy}

\vspace{2pt}
\begin{tabular}{lrrrrrr}
\toprule
Head function & Char. & Typo & Token & Word & Synonym & Shuffle \\
\midrule
Previous token & $-0.217$ & $-0.221$ & $-0.467$ & $-0.359$ & $-0.195$ & $-0.670$ \\
Duplicate      & $0.016$ & $-0.005$ & $0.218$ & $0.182$ & $0.169$ & $0.559$ \\
Induction      & $0.271$ & $0.312$ & $0.571$ & $0.503$ & $0.329$ & $0.609$ \\
Copying        & $0.097$ & $0.045$ & $0.134$ & $0.252$ & $0.331$ & $0.248$ \\
\bottomrule
\end{tabular}
\end{table}

\begin{table}[t]
\centering
\small
\renewcommand{\arraystretch}{1.15}
\setlength{\tabcolsep}{5.5pt}

\caption{Spearman correlations between head-function scores and percentage changes in the patched model's output metrics under 30\% perturbation on GPT-2. Copying entries use the supplied corrected scores. These descriptive correlations pool repeated observations from all 144 GPT-2 heads. No seed-level uncertainty or significance claim is made. Head-mean and controlled analyses are reported in Appendix~\ref{headcontrols}.}
\label{tab:patching}

\vspace{2pt}
\begin{tabular}{lrrrr}
\toprule
& \multicolumn{2}{c}{Patched $\Delta\%\,\mathrm{NLL}$}
& \multicolumn{2}{c}{Patched $\Delta\%\,\mathrm{OutDiv}$} \\
\cmidrule(lr){2-3}
\cmidrule(lr){4-5}
Head function & Token & Shuffle & Token & Shuffle \\
\midrule
Previous token & $0.147$ & $-0.011$ & $0.218$ & $-0.036$ \\
Duplicate      & $0.163$ & $0.247$  & $0.026$ & $0.057$  \\
Induction      & $-0.153$ & $-0.017$ & $-0.173$ & $0.053$ \\
Copying        & $0.736$ & $0.603$  & $0.661$ & $0.344$ \\
\bottomrule
\end{tabular}
\end{table}

We relate the response measures in Appendix~\ref{attnentnormdelta} and Section~\ref{patchingdesc} to the four head-function scores. In Table~\ref{tab:attention_entropy}, induction score has the largest-magnitude positive correlation with perturbed-minus-clean entropy change for most perturbations ($0.271$ to $0.609$). Under shuffling, previous-token score has the strongest negative correlation ($-0.670$), while duplicate and induction scores are strongly positive. The layerwise plots in Appendix~\ref{headcontrols} reveal substantial variation across layers, including a negative late-layer previous-token association with entropy change. These associations describe attention redistribution and do not establish a causal role for a head type.

For the token-count-preserving perturbations used in patching, copying score has the strongest pooled association with recovery (Table~\ref{tab:patching}): correlations using the supplied corrected copying scores are $0.736$ and $0.603$ for NLL recovery under token substitution and shuffling, and $0.661$ and $0.344$ for output-divergence recovery. These pooled estimates do not account for repeated observations or layer and head-scale differences. In a separate analysis using one mean per head, copying score remains positively associated with recovery after accounting for layer and OV norm: partial rank correlations are $0.733$ and $0.505$ for NLL, and $0.782$ and $0.437$ for output divergence, under token substitution and shuffling, respectively (Appendix~\ref{headcontrols}). These controls support a conditional association in this setting, but do not establish a copying-specific causal mechanism.

\subsection{RQ3: Cross-model and cross-scale generalization}

\begin{table}[t]
\centering
\small
\renewcommand{\arraystretch}{1.12}
\setlength{\tabcolsep}{3.5pt}

\caption{Family differences in baseline-adjusted NLL change and output divergence. GPT comprises the four GPT-2 checkpoints; Qwen comprises Qwen2.5--0.5B and Qwen2.5--1.5B. NLL entries are point estimates obtained by subtracting the reported clean family gap ($0.592$) from the reported perturbed family gap; values are approximate because the source means are rounded. Adjusted NLL uncertainty is not reported because paired sequence-level data are needed. Output-divergence entries retain the original mean$_{\mathrm{SD}}$ across data samples. Negative NLL entries indicate a smaller increase relative to clean inputs for GPT.}
\label{tab:cross_model_behavioral}

\vspace{2pt}
\begin{tabular}{lrrrrrr}
\toprule
Perturbation rate & Char. & Typo & Token & Word & Synonym & Shuffle \\
\midrule
\multicolumn{7}{l}{\textit{Baseline-adjusted NLL change difference}} \\
0\% & $0.000$ & $0.000$ & $0.000$ & $0.000$ & $0.000$ & $0.000$ \\
5\% & $0.052$ & $0.096$ & $-0.154$ & $-0.053$ & $-0.020$ & $0.108$ \\
30\% & $-0.695$ & $-0.519$ & $-0.860$ & $-0.135$ & $-0.042$ & $0.299$ \\
50\% & $-0.962$ & $-0.822$ & $-1.302$ & $-0.236$ & $-0.070$ & $0.263$ \\
\addlinespace[3pt]
\midrule
\multicolumn{7}{l}{\textit{Output divergence}} \\
0\%  & $0.000_{0.000}$ & $0.000_{0.000}$ & $0.000_{0.000}$ & $0.000_{0.000}$ & $0.000_{0.000}$ & $0.000_{0.000}$ \\
5\%  & $0.043_{0.008}$ & $0.043_{0.008}$ & $0.002_{0.008}$ & $-0.017_{0.006}$ & $-0.033_{0.006}$ & $0.030_{0.008}$ \\
30\% & $0.048_{0.005}$ & $0.041_{0.005}$ & $0.008_{0.006}$ & $0.018_{0.005}$ & $0.010_{0.008}$ & $0.071_{0.010}$ \\
50\% & $0.032_{0.005}$ & $0.024_{0.005}$ & $-0.012_{0.005}$ & $0.008_{0.004}$ & $0.023_{0.007}$ & $0.059_{0.009}$ \\
\bottomrule
\end{tabular}
\end{table}
We compare the four GPT-2 checkpoints with the two Qwen2.5 checkpoints using NLL, output divergence, and intrinsic-dimension change. We omit cross-model CKA comparisons because layer counts differ and position-wise interpretation is unreliable for perturbations that alter tokenization.

Table~\ref{tab:cross_model_behavioral} adjusts NLL family differences for the clean baseline. For family $F$, let $\overline{\mathrm{NLL}}_F(p)$ average over the tested checkpoints and evaluation sequences. We report
\begin{align*}
D_{\Delta\mathrm{NLL}}(p)
&= \bigl[\overline{\mathrm{NLL}}_{\mathrm{GPT}}(p)-\overline{\mathrm{NLL}}_{\mathrm{GPT}}(0)\bigr] \\
&\quad-\bigl[\overline{\mathrm{NLL}}_{\mathrm{Qwen}}(p)-\overline{\mathrm{NLL}}_{\mathrm{Qwen}}(0)\bigr].
\end{align*}
The reported clean gap is $0.592$. Subtracting it reverses the token-substitution gap already at 5\% ($0.438-0.592=-0.154$); the adjusted gaps at 30\% and 50\% are $-0.860$ and $-1.302$. Word and synonym substitution also have negative adjusted NLL gaps at all reported nonzero rates, whereas shuffling has positive gaps. Output-divergence differences are generally small, with shuffling showing the largest positive gap at 30\% and 50\%. These point estimates describe the tested checkpoints; they do not establish statistical significance or an architectural cause. Baseline adjustment does not resolve cross-tokenizer comparability of per-token NLL or unequal model-size composition of the families.

\begin{figure}
    \centering
    \includegraphics[width=1\linewidth]{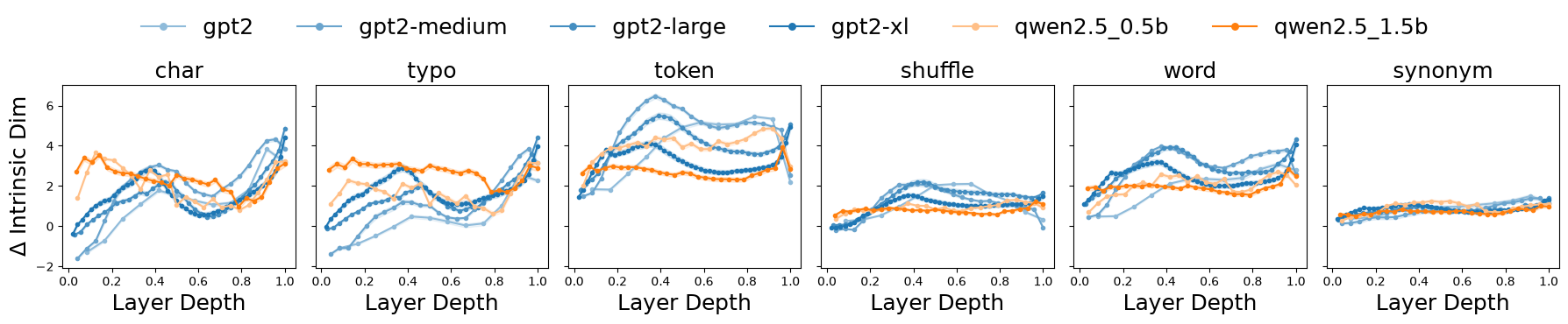}
    \caption{Two-NN Intrinsic dimension estimate for all perturbation types, 30\% perturbation, across GPT-2 and Qwen2.5 checkpoints.}
    \label{fig:cross_model_intrinsic_dim}
\end{figure}

Figure~\ref{fig:cross_model_intrinsic_dim} compares intrinsic-dimension changes at 30\% perturbation. Character and typo perturbations show broadly similar later-layer trends, with larger early-layer increases in Qwen. These profiles do not isolate sensitivity to retokenization. GPT checkpoints show larger intrinsic-dimension changes for token substitution, shuffling, and word substitution, while synonym substitution produces comparatively small changes. The geometric ordering need not match behavioral ordering: for token and word substitution, the baseline-adjusted NLL gap is negative despite the larger GPT geometric changes. Accordingly, these profiles support metric-dependent descriptions rather than a single family-level robustness ranking.

\subsection{RQ4: Adversarial perturbation and worst-case behavior}

\begin{figure}[tbp]
    \centering
    \makebox[\linewidth][c]{%
        \includegraphics[width=0.485\linewidth]{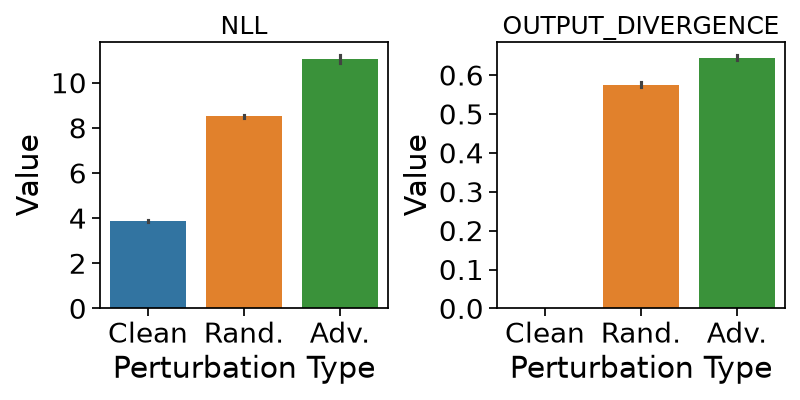}%
        \hspace{0.015\linewidth}%
        \includegraphics[width=0.500\linewidth]{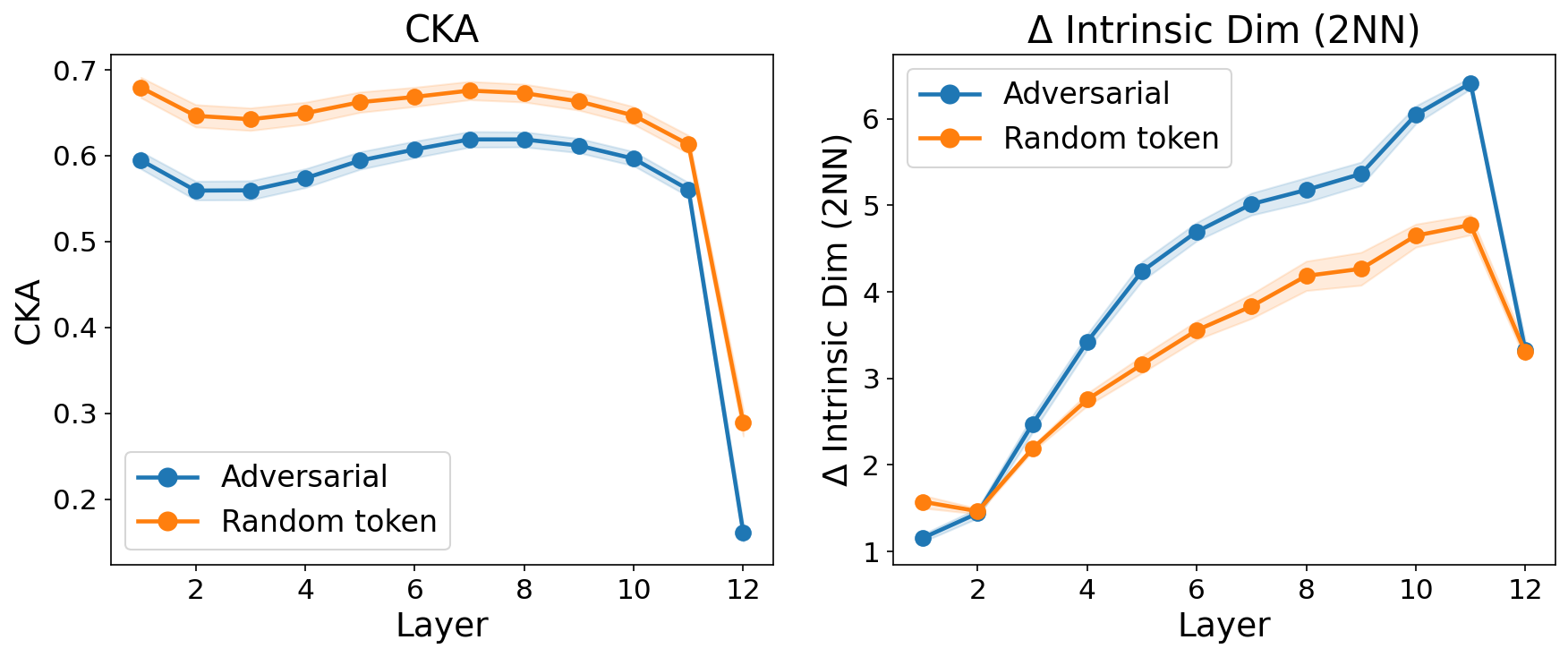}%
    }
    \caption{Adversarial versus random token substitution on GPT-2 at 30\% perturbation strength. The left panel shows behavioral differences, while the right panel shows internal representational differences.}
    \label{fig:adv-comparison}
\end{figure}

\begin{table*}[t]
\caption{Negative log-likelihood (NLL) and output divergence under adversarial HotFlip perturbations and random token substitutions for six language models at perturbation strengths of 5\% and 30\%. Entries report the mean and standard deviation across evaluation sequences.}
\label{tab:adv_vs_token_nll_by_pct}
\vspace{4pt}
\centering
\small
\setlength{\tabcolsep}{4.5pt}
\renewcommand{\arraystretch}{1.18}

\begin{tabular}{lrrrrrrrr}
\toprule
& \multicolumn{4}{c}{Negative log-likelihood}
& \multicolumn{4}{c}{Output divergence} \\
\cmidrule(lr){2-5}
\cmidrule(lr){6-9}
& \multicolumn{2}{c}{5\%} & \multicolumn{2}{c}{30\%}
& \multicolumn{2}{c}{5\%} & \multicolumn{2}{c}{30\%} \\
\cmidrule(lr){2-3}
\cmidrule(lr){4-5}
\cmidrule(lr){6-7}
\cmidrule(lr){8-9}
Model & Adv. & Rand. & Adv. & Rand. & Adv. & Rand. & Adv. & Rand. \\
\midrule
GPT-2
& $5.38_{0.50}$ & $4.81_{0.58}$ & $11.03_{1.83}$ & $8.52_{0.50}$
& $0.25_{0.08}$ & $0.18_{0.09}$ & $0.64_{0.05}$ & $0.57_{0.05}$ \\
GPT-2 Medium
& $5.07_{0.53}$ & $4.54_{0.68}$ & $9.87_{0.70}$ & $8.41_{0.55}$
& $0.25_{0.08}$ & $0.18_{0.10}$ & $0.63_{0.04}$ & $0.58_{0.05}$ \\
GPT-2 Large
& $5.03_{0.46}$ & $4.38_{0.56}$ & $10.68_{0.87}$ & $8.35_{0.50}$
& $0.24_{0.07}$ & $0.18_{0.09}$ & $0.61_{0.05}$ & $0.58_{0.05}$ \\
GPT-2 XL
& $4.89_{0.43}$ & $4.29_{0.54}$ & $10.34_{0.72}$ & $8.29_{0.51}$
& $0.24_{0.07}$ & $0.18_{0.09}$ & $0.62_{0.05}$ & $0.58_{0.05}$ \\
Qwen2.5--0.5B
& $4.81_{0.46}$ & $4.22_{0.62}$ & $10.09_{0.47}$ & $8.87_{0.62}$
& $0.26_{0.06}$ & $0.18_{0.08}$ & $0.63_{0.04}$ & $0.58_{0.05}$ \\
Qwen2.5--1.5B
& $4.48_{0.45}$ & $3.92_{0.62}$ & $9.80_{0.50}$ & $8.61_{0.63}$
& $0.24_{0.06}$ & $0.18_{0.08}$ & $0.62_{0.04}$ & $0.57_{0.05}$ \\
\bottomrule
\end{tabular}
\end{table*}

We compare HotFlip with random token substitutions matched at the edited positions. Figure~\ref{fig:adv-comparison} reports the full behavioral and internal comparison for GPT-2 at 30\% perturbation. HotFlip yields higher NLL and output divergence, lower CKA to clean activations, and a larger increase in estimated intrinsic dimension than the matched random control. These results show that random substitution at the same positions does not reproduce the attack's GPT-2 response profile.

Table~\ref{tab:adv_vs_token_nll_by_pct} extends the behavioral comparison to all six checkpoints at 5\% and 30\% perturbation. HotFlip has higher mean NLL and output divergence than its matched random control in every reported cell. This consistency applies to the tested checkpoints and rates; the internal metrics in Figure~\ref{fig:adv-comparison} remain specific to GPT-2 at 30\%.

\section{Discussion}
\paragraph{Surface and internal measures can dissociate.}
Different metrics yield different descriptions of perturbation effects. Within RQ1, token substitution produces output divergence comparable to word substitution but higher NLL. Within RQ2, copying scores have stronger pooled correlations with patching recovery than with entropy change, and the separate head-mean association remains positive after accounting for layer and OV norm (Appendix~\ref{headcontrols}); general head influence remains a possible explanation. Within RQ3, token substitution yields small output-divergence gaps, a negative baseline-adjusted NLL gap, and larger GPT intrinsic-dimension changes. These observations do not support a uniform family-level robustness ordering. Within RQ4, HotFlip yields higher NLL and output divergence, lower CKA, and larger intrinsic-dimension increases than its matched random control in GPT-2 at 30\%. Behavioral metrics therefore provide an incomplete description of measured internal changes.

\paragraph{Limitations.}
CKA for character, typo, word, and synonym perturbations combines representation change with retokenization and positional mismatch (Section~\ref{cka}); truncation ensures equal shape, not semantic alignment. RQ2 includes descriptive pooled estimates and separate head-mean analyses controlling for layer and OV norm. These associations do not isolate a copying-specific causal mechanism. The bootstrap intervals in Appendix~\ref{headcontrols} are conditional on the observed head means and treat heads as resampling units; they do not quantify variation across perturbation seeds, sequences, or model checkpoints. The OV-norm covariate is the Frobenius norm $\|W_v^{(h)}W_o^{(h)}\|_F$, as defined in the analysis implementation. Corrected copying scores are available only for token substitution and shuffling. RQ3's adjusted NLL means are calculated from rounded summary values; paired sequence-level data are required to quantify their uncertainty. Different tokenizers, model sizes, and training histories also limit cross-family interpretation. Architectural explanations involving grouped-query attention, rotary positional embeddings, or vocabulary size remain untested. All analyses use WikiText-2 and two model families; broader generality requires additional domains and models.

\paragraph{Implications.}
Complementary metrics can expose discrepancies that a single measure would conceal. In RQ4, the accepted-position-matched random control supports attributing the observed GPT-2 response differences to replacement selection beyond edit position and rate alone. Agreement across metrics provides corroborating evidence under this design, but does not independently establish a unique mechanism or exclude shared measurement confounds. Metric agreement and disagreement should both motivate checks of alignment, baseline differences, and alternative explanations before results are generalized.

\section{Conclusion}
We find distinguishable perturbation profiles in language-model behavior, hidden-state geometry, and attention-head response measures. Their ordering depends on the metric and is only partly consistent across the tested checkpoints. Pooled copying-score correlations do not establish a copying-specific recovery mechanism, and baseline-adjusted NLL differences do not support a uniform cross-family robustness ranking. In the matched adversarial comparison, agreement across behavioral and internal metrics provides corroborating evidence of disruption under the tested conditions rather than proof of a unique underlying mechanism.

Future work should quantify uncertainty in paired baseline-adjusted comparisons, test alignment-aware representation measures, and extend the head-level analyses to other models and domains. Architectural explanations for family differences remain open, and the matched adversarial design can be extended to other perturbation types. More broadly, robustness should be evaluated using complementary measurements with explicit attention to their assumptions and confounds.

\bibliographystyle{plainnat}
\bibliography{references}

\newpage

\appendix

\section{Metrics}

\subsection{Output Metrics}\label{output_metrics}

\subsubsection{NLL}
Sequence-level NLL on an output sequence $y$ (either clean or perturbed) measures how well the model predicts the next token under a perturbed input sequence. Lower NLL indicates that the model assigns higher probability to the observed continuation.
\begin{align*}
    \mathrm{NLL}(y)=-\frac{1}{T}\sum_{t=1}^{T}\log p(y_t\mid y_{<t})
\end{align*}

\subsubsection{Output Divergence}
To compare generated outputs under clean and perturbed inputs, we computed normalized Levenshtein distance between generated sequences $y$ (clean) and $\hat{y}$ (perturbed):

\begin{align*}
    \mathrm{OutputDivergence}(y,\hat y)=\frac{\mathrm{EditDistance}(y,\hat y)}{\max(|y|,|\hat y|)}
\end{align*}

where EditDistance is the Levenshtein edit distance. This measures how much the generated continuation changes under perturbation.

\subsection{Intrinsic Dimensions}\label{intrinsic_dim_metrics}

\subsubsection{2-Nearest Neighbors}
For each data point $i$, in this instance an individual token's activation on layer $l$ (like in CKA, $1 \leq l \leq n_{layer}$), we compute distances $r_1^i$ and $r_2^i$ between the point and its two nearest neighbors within the batch. Defining $$\mu_i = \frac{r_2^i}{r_1^i},$$ we use the maximum likelihood estimator to approximate the intrinsic dimension $d$: 
\begin{align*}
    d = \frac{N}{\sum\ln\mu_i}
\end{align*}
where $N$ is the number of total points in the batch.

\subsection{Attention Function Scores}\label{attn_func_scores}
\subsubsection{Previous-Token Score}
We report this score as an average of all places in the attention matrix where position $i$ attends to position $i-1$. The attention matrix $A$ used by the previous token score is computed using clean inputs $x$ ($\alpha = h(x)$), and the overall score is averaged over all inputs:
\begin{align*}
    \mathrm{Previous\_Token}(h) = \mathbb{E}_x\left[\frac{1}{L-1}\sum_{i=1}^{L-1}\alpha_{i, i-1}\right]. 
\end{align*}
\subsubsection{Duplicate Score}
In order to isolate the duplicate detection signal from previous-token signal and general semantic associations, we construct an input probe by repeating a randomized sequence of $r$ of $L$ tokens twice. Then, we evaluate the average attention values in $\alpha = h(rr)$ where positions in the second sequence attend to their corresponding positions in the first sequence:
\begin{align*}
    \mathrm{Duplicate}(h) = \frac{1}{L}\sum_{i=L+1}^{2L}\alpha_{i, i-L}.
\end{align*}
\subsubsection{Induction Score}

We use the same repeated random token probe as the duplicate score, but examine the attention between the positions in the repeated sequence and the positions immediately following their counterparts in the earlier sequence.
\begin{align*}
    \mathrm{Induction}(h) = \frac{1}{L-1}\sum_{i=L+1}^{2L-1}\alpha_{i, i-L+1}.
\end{align*}

\subsection{Attention Entropy Norm Delta}\label{attnentnormdelta}
Given a clean input $x$ and a perturbed input $x'$, we obtain attention matrices $\alpha = h(x)$ and $\alpha'=h(x')$. In our experiments, we construct $x'$ under 30\% perturbation for all perturbation types. Defining entropy at position $i$ as
\begin{align*}
    \widehat{H}(\alpha)
    = \frac{1}{L-1}\sum_{i=1}^{L-1}
      \frac{H_i(\alpha)}{\log(i+1)}.
\end{align*}

we take the normed entropy over all positions excluding the first (since the first position does not have a target attention distribution to draw from). The normed entropy divides the total entropy at each location by the maximum possible entropy, to ensure that all entropies are evaluated at comparable scale:

Finally, the \emph{entropy norm delta} over all inputs is given by
\begin{align*}
    \Delta{H} = \mathbb{E}_x\left[\hat{H}(\alpha') - \hat{H}(\alpha)\right].
\end{align*}
Positive $\Delta H$ indicates more diffuse perturbed attention; negative values indicate greater concentration. Tables and plots use this perturbed-minus-clean convention.

\section{Adversarial Shortlist Selection}\label{advshortlist}

\begin{figure}
    \centering
    \includegraphics[width=0.6\linewidth]{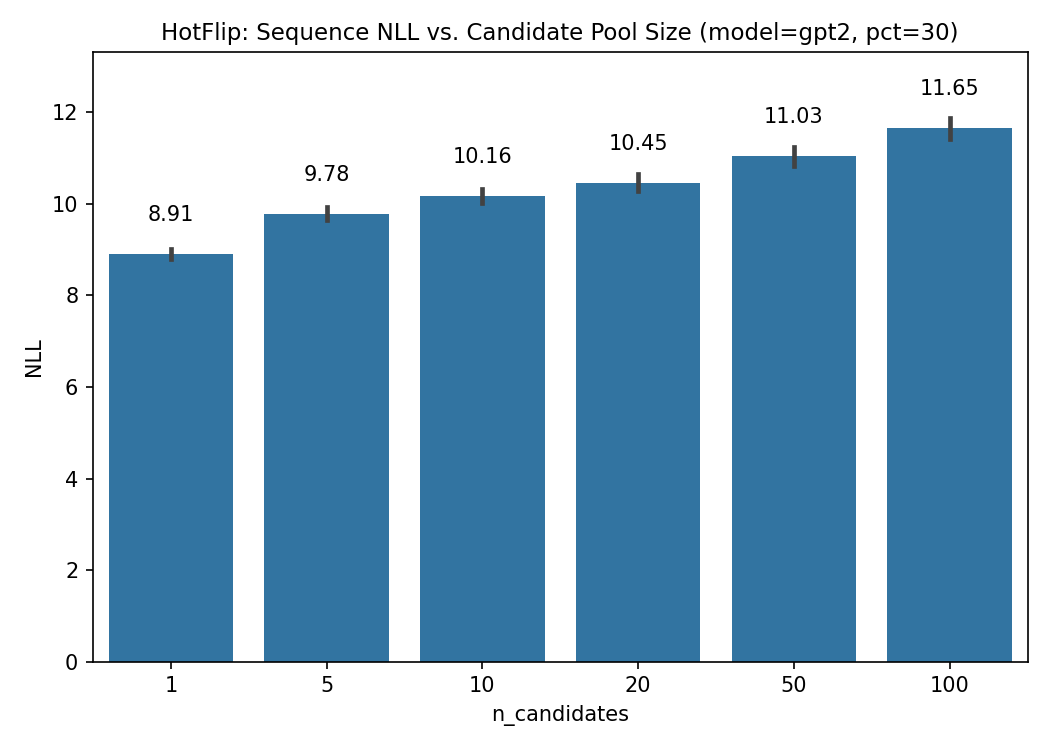}
    \caption{Sequence NLL as a function of HotFlip's gradient-shortlist size ($n$\_candidates), on GPT-2 at 30\% perturbation strength
    ($n=300$). }
    \label{fig:n_candiadates_nll}
\end{figure}

We fix $|S|=50$ based on an ablation over $|S| \in \{1, 5, 10, 20, 50, 100\}$ on GPT-2 at $p=30\%$ (Figure~\ref{fig:n_candiadates_nll}). Mean NLL rises from 8.91 ($|S|{=}1$) to 11.65 ($|S|{=}100$), while the marginal gain per additional candidate declines from 0.22 between $|S|{=}1$--$5$ to 0.01 between $|S|{=}50$--$100$. We select $|S|=50$ as a practical balance between attack strength and compute cost rather than as a loss-maximizing choice.

\section{Head-level associations and controls}\label{headcontrols}
\paragraph{Aggregation and layer numbering.}
Each supplied condition CSV contains 2,304 rows: 16 observations for each of the 144 GPT-2 heads. These records do not identify perturbation seeds, so we do not treat the 16 observations as independent seed runs. For the analyses below, we take arithmetic means within each layer--head pair. Layerwise Spearman correlations use the 12 distinct heads in each layer. Layers 0--11 denote the first through twelfth transformer attention blocks, not the embedding layer.

\paragraph{Exploratory uncertainty.}
For each layerwise estimate, we draw 2,000 paired bootstrap samples of the 12 heads with replacement, keeping each head's score and outcome together, and report the 2.5th and 97.5th percentiles. Undefined constant-score resamples are excluded; a layerwise band is omitted when fewer than 90\% of draws are valid. For the across-head partial rank correlations, we resample 12 heads within each of the 12 layers, rerank the variables and refit the residualization in each of 2,000 draws. These exploratory intervals quantify variation under head resampling conditional on the observed head means. They are not seed-level or sequence-level intervals, are not adjusted for multiple comparisons, and do not establish a copying-specific mechanism.

\begin{figure}[t]
    \centering
    {\small (a) Attention entropy change}\par\smallskip
    \includegraphics[width=\linewidth]{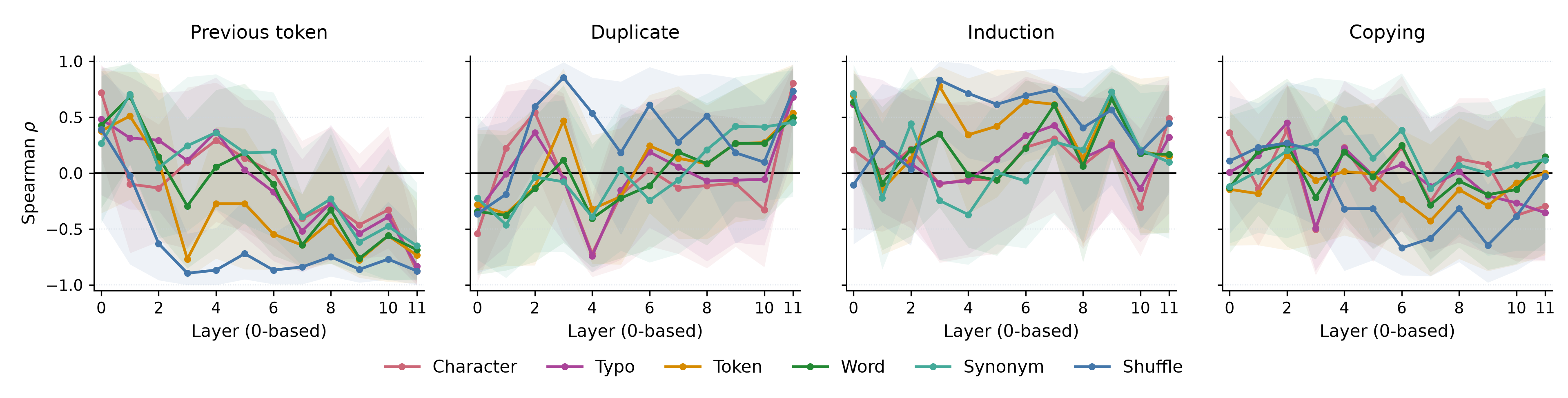}\par\medskip
    {\small (b) NLL patching recovery}\par\smallskip
    \includegraphics[width=\linewidth]{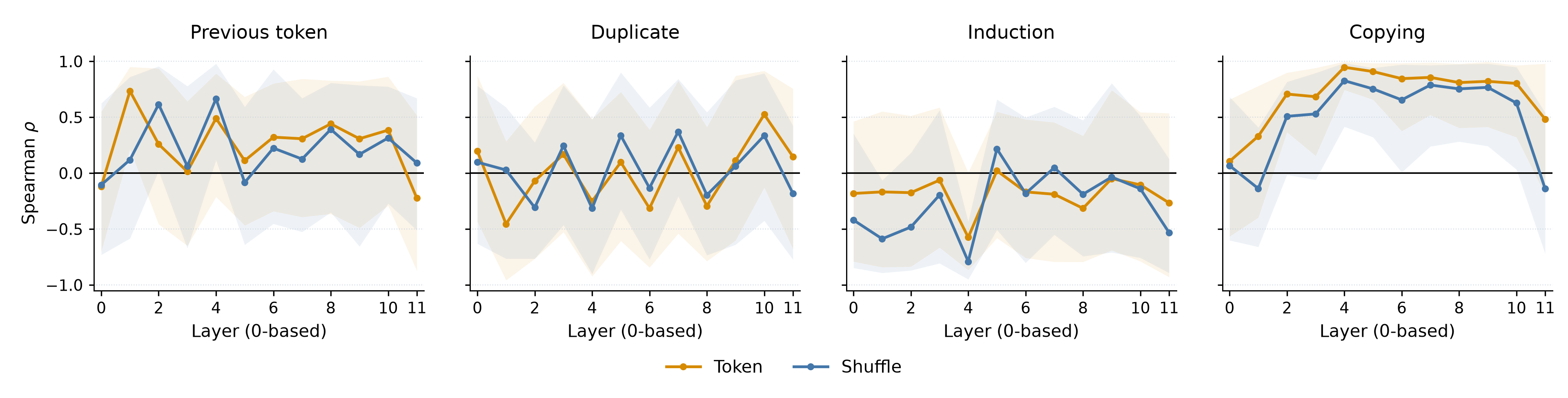}\par\medskip
    {\small (c) Output-divergence patching recovery}\par\smallskip
    \includegraphics[width=\linewidth]{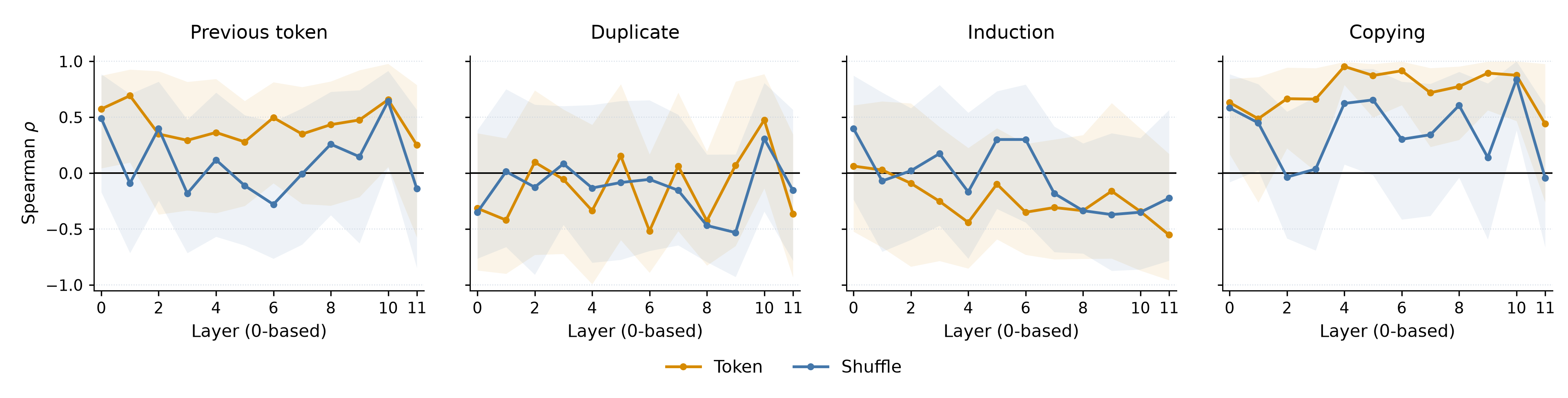}
    \caption{Layerwise Spearman correlations between head-function scores and (a) normalized attention entropy change, (b) NLL patching recovery, and (c) output-divergence patching recovery at 30\% perturbation. Entropy is perturbed minus clean; patching uses token substitution and shuffling. Copying panels use supplied corrected scores for token and shuffle and original scores for the other four entropy conditions. Points use the 12 head means per layer; shading shows exploratory 95\% paired head-bootstrap intervals.}
    \label{fig:perlayer_head_associations}
    \label{fig:perlayer_attn_entropy}
    \label{fig:perlayer_nll_improvement}
    \label{fig:perlayer_outdiv_improvement}
\end{figure}

\begin{table}[t]
\centering
\small
\renewcommand{\arraystretch}{1.15}
\setlength{\tabcolsep}{5.5pt}

\caption{Supplied corrected copying-score associations with patching recovery across 144 head means. Partial rank correlations account for layer indicators and ranked OV norm; intervals refer to these partial correlations and use 2,000 head-bootstrap samples stratified by layer. Aggregation differs from Table~\ref{tab:patching}.}
\label{tab:copying_controls}

\vspace{2pt}
    \begin{tabular}{llrrl}
        \toprule
        Perturbation & Recovery metric & Spearman $\rho$ & Partial rank $r$ & 95\% interval \\
        \midrule
        Token & NLL & 0.755 & 0.733 & [0.618, 0.822] \\
        Token & Output divergence & 0.768 & 0.782 & [0.692, 0.845] \\
        Shuffle & NLL & 0.640 & 0.505 & [0.356, 0.643] \\
        Shuffle & Output divergence & 0.572 & 0.437 & [0.285, 0.568] \\
        \bottomrule
    \end{tabular}
\end{table}

\paragraph{Scope of the OV control.}
We merge the corrected copying scores and OV-norm values with the recovery records by layer and head; all 144 heads match in each condition. OV norm is computed as $\|W_v^{(h)}W_o^{(h)}\|_F$. The copying-score formula is unchanged in the supplied corrected analysis. Positive partial correlations show that accounting for layer and OV norm does not remove the observed recovery association in this analysis. This conditional association does not establish a copying-specific causal mechanism.

\end{document}